\def\year{2021}\relax
\documentclass[letterpaper]{article} % DO NOT CHANGE THIS
\usepackage{aaai21}  % DO NOT CHANGE THIS
\usepackage{times}  % DO NOT CHANGE THIS
\usepackage{helvet} % DO NOT CHANGE THIS
\usepackage{courier}  % DO NOT CHANGE THIS
\usepackage[hyphens]{url}  % DO NOT CHANGE THIS
\usepackage{graphicx} % DO NOT CHANGE THIS
\usepackage{natbib}  % DO NOT CHANGE THIS AND DO NOT ADD ANY OPTIONS TO IT
\usepackage{caption} % DO NOT CHANGE THIS AND DO NOT ADD ANY OPTIONS TO IT
\usepackage{amsfonts}       % blackboard math symbols
\usepackage{nicefrac}       % compact symbols for 1/2, etc.
\usepackage{microtype}      % microtypography
\usepackage{amsmath}
\usepackage{multicol}
\usepackage[displaymath, mathlines]{lineno}
\usepackage{subfigure}
\usepackage{commath}
\usepackage{amssymb}
\usepackage{nccmath}
\usepackage{float}
\usepackage{bm}
\usepackage{dsfont}
\usepackage{mdframed}
\usepackage{multirow}
\usepackage{tabularx}
\usepackage{tabulary}
\usepackage{amsthm}
\usepackage{tikz}
\usepackage{rotating}
\usepackage{mathtools}% Loads amsmath
\usepackage{booktabs}
\usepackage{adjustbox}
\usepackage{algorithm}
\usepackage{algpseudocode}
\usepackage{tabularx}
\usepackage[vlines]{tabularht}
\usepackage{subfiles} % Best loaded last in the preamble
\usepackage{lineno}
\usepackage{wrapfig}

\iftrue
\usepackage{lastpage}
\usepackage{fancyhdr}
\fi
\usepackage{hyperref}

\title{The Target Polish: A New Approach to Outlier-Resistant Non-Negative Matrix Factorization}

\author{
    Paul Fogel,\textsuperscript{\rm 1} Christophe Geissler,\textsuperscript{\rm 1} George Luta\textsuperscript{\rm 2, \thanks{To whom the correspondence should be addressed.}}\\
}
\affiliations{
    \small
    \textsuperscript{\rm 1}Data Services, Forvis Mazars, Levallois, France\\
    \textsuperscript{\rm 2} Department of Biostatistics, Bioinformatics and Biomathematics, Georgetown University Medical Center, Washington, DC, USA \\
    paul.fogel@forvismazars.com,
    christophe.geissler@forvismazars.com,
    george.luta@georgetown.edu
}

\begin{document}

\maketitle
\begin{abstract}
This paper introduces the "Target Polish," a robust and computationally efficient framework for Non-Negative Matrix Factorization (NMF). Although conventional weighted NMF approaches are resistant to outliers, they converge slowly due to the use of multiplicative updates to minimize the objective criterion. In contrast, the Target Polish approach remains compatible with the Fast-HALS algorithm, which is renowned for its speed, by adaptively "polishing" the data with a weighted median-based transformation. This innovation provides outlier resistance while maintaining the highly efficient additive update structure of Fast-HALS. Empirical evaluations using image datasets corrupted with structured (block) and unstructured (salt) noise demonstrate that the Target Polish approach matches or exceeds the accuracy of state-of-the-art robust NMF methods while reducing computational time by an order of magnitude in the studied scenarios. \newline

\end{abstract}

\section{Introduction}

Non-negative Matrix Factorization (NMF) decomposes a non-negative matrix \(X \in \mathbb{R}_+^{m \times n}\) into two non-negative \textit{factoring matrices} \(W \in \mathbb{R}_+^{m \times r}\) and \(H \in \mathbb{R}_+^{n \times r}\), such that:
\begin{equation} \label{eq:nmf}
    X\approx WH^T.
\end{equation}
Typically, the rank \(r\) is selected such that  \(r \ll m,n\). A common optimization objective for NMF is the minimization of the Frobenius norm difference between \(X\) and \(WH^T\):

\begin{equation} \label{eq:cost_nmf}
    J_{\text{NMF}} = \sum_{i,j} \left( X_{ij} - (WH^T)_{ij} \right)^2.
\end{equation}

In image analysis, which is the area we focus on in the current work, it is common to represent each image as a column in a matrix, with each row corresponding to a specific pixel. In this case, since the features are pixel intensities, they naturally share the same scale. However, in more general scenarios, it is recommended to scale the features before applying NMF to ensure that each one contributes equally, regardless of its original magnitude. \newline
 
Thanks to the non-negativity constraints, NMF factoring matrices are typically sparse and interpretable \cite{Lee1999LearningTP}. These unique characteristics have played a key role in its early adoption within the field of dimension reduction techniques \cite{Paatero1994}. NMF is widely used in image processing, text mining, and bioinformatics, as it helps uncover hidden data structures while ensuring interpretability \cite{Guillamet2002}, \cite{Berry2007}, \cite{Devarajan2008}. Its extension to multi-dimensional arrays, Non-negative Tensor Factorization (NTF), applies the same principles while going beyond matrices \cite{Cichocki2009}. \newline

Among the numerous related algorithms developed  for NMF so far, Fast-HALS (Hierarchical Alternating Least Squares) is considered one of the most powerful in terms of computational performance \cite{Cichocki2009FastLA}. Its remarkable convergence properties have been recently studied \cite{Hou2024ConvergenceOA}. To better understand these properties, the following provides a concise review of the fundamental mathematics underlying HALS and Fast-HALS. For simplicity, the formulation focuses on matrices, though it can be extended to tensors of any dimension. \newline

Consider \(X_k =  (X - W H^T + w_k h_k^T)\), which represents the sum of the portion of the factorization explained by the \(k^{th}\) component (or \textit{part}) and the residual. To update \(H\), HALS updates each component \(h_k\) by projecting the matrix \(X_k \) on \(w_k\) using:
\begin{equation} \label{eq:hals_1}
   h_k \leftarrow \left[(X_k^T w_k)/{w_k^Tw_k}\right]_+.
\end{equation}
Fast-HALS follows different update rules: Assuming \(\|w_k\|_2 = 1\), equation \ref{eq:hals_1} can be rewritten as:
\begin{equation} \label{eq:hals_2}
    h_k \leftarrow \left[(X - WH^T + w_k h_k^T)^T w_k\right]_+.
\end{equation}
Using the associativity of matrix multiplication, Fast-HALS further simplifies the equation \ref{eq:hals_2}:
\begin{equation} \label{eq:fast_hals}
    h_k \leftarrow \left[ h_k + \left[ X^T W \right]_k - H \left[ W^T W \right]_k \right]_+.
\end{equation}
This method eliminates the need to explicitly compute each component \(X_k\), substituting it with a single matrix multiplication \( X^T W\). Moreover, the additive structure of the update rule operates within a \((n,r)\)-dimensional space rather than a \((n,m)\)-dimensional space, yielding a reduction factor of \(r/m\). Since \(r \ll m\),  this significantly enhances computational efficiency. \newline

However, like most algorithms using Frobenius norm-based objectives, Fast-HALS is sensitive to outliers. Incorporating resistance to outliers in the design of algorithms commonly implies that ”the least squares criterion is replaced by a weighted least squares criterion, where the weight function is chosen in order to give less weight to ”discrepant” observations (in the sense only of fitting the model less well)” \cite{Green1984IterativelyRL}. Weighted NMF incorporates a weight matrix \(G_{ij}\) to mitigate the impact of outliers, modifying the optimization objective:
\begin{equation} \label{eq:weighted_nmf}
    J_{\text{Weighted NMF}} = \sum_{i,j} G _{ij} \left( X_{ij} - (WH^T)_{ij} \right)^2.
\end{equation}

A number of weighting schemes have been proposed, such as the popular Correntropy Induced Metric (CIM) approach and the Huber approach \cite{Du2012RobustNM}, \cite{Wang2019SparseGR}.
\begin{itemize}
    \item CIM-NMF adopts an exponential function:
\begin{equation} \label{eq:cim}
    G _{ij} = \exp \left( -\frac{(X_{ij} - (WH^T)_{ij})^2}{\sigma^2} \right)
\end{equation}
where \(\sigma^2\) is the variance of the matrix entries. This suppresses large deviations and effectively handles extreme outliers, particularly in image processing applications.

\item Huber-NMF balances robustness to outliers and accuracy with a weight function:
\begin{equation} \label{eq:huber}
    G _{ij} = 
    \begin{cases}
        1, & |X_{ij} - (WH^T)_{ij}| \leq \delta \\
        \frac{\delta}{|X_{ij} - (WH^T)_{ij}|}, & \text{otherwise}.
    \end{cases}
\end{equation}
Here, \(\delta\) is the median absolute error between \(X\) and \(WH^T\), ensuring that well-approximated entries retain full weight while those with larger deviations are scaled.
 
\end{itemize}

Unfortunately, the associativity of matrix multiplication, which is crucial for the Fast-HALS update rules, no longer holds when weight matrices are introduced. As a result, weighted NMF algorithms rely on multiplicative update rules, which converge sub-linearly to asymptotic stability \cite{Badeau2011StabilityAO} and generally are slower than Fast-HALS.\newline

In this work, we introduce a novel approach: the ”Target Polish”, which allows us to apply Fast-HALS rules and take full advantage of their computational power, while being resistant to outliers.

\section{Materials and methods}
\subsection{Materials}
\begin{figure}
    \centering
    \includegraphics[width=0.5\linewidth]{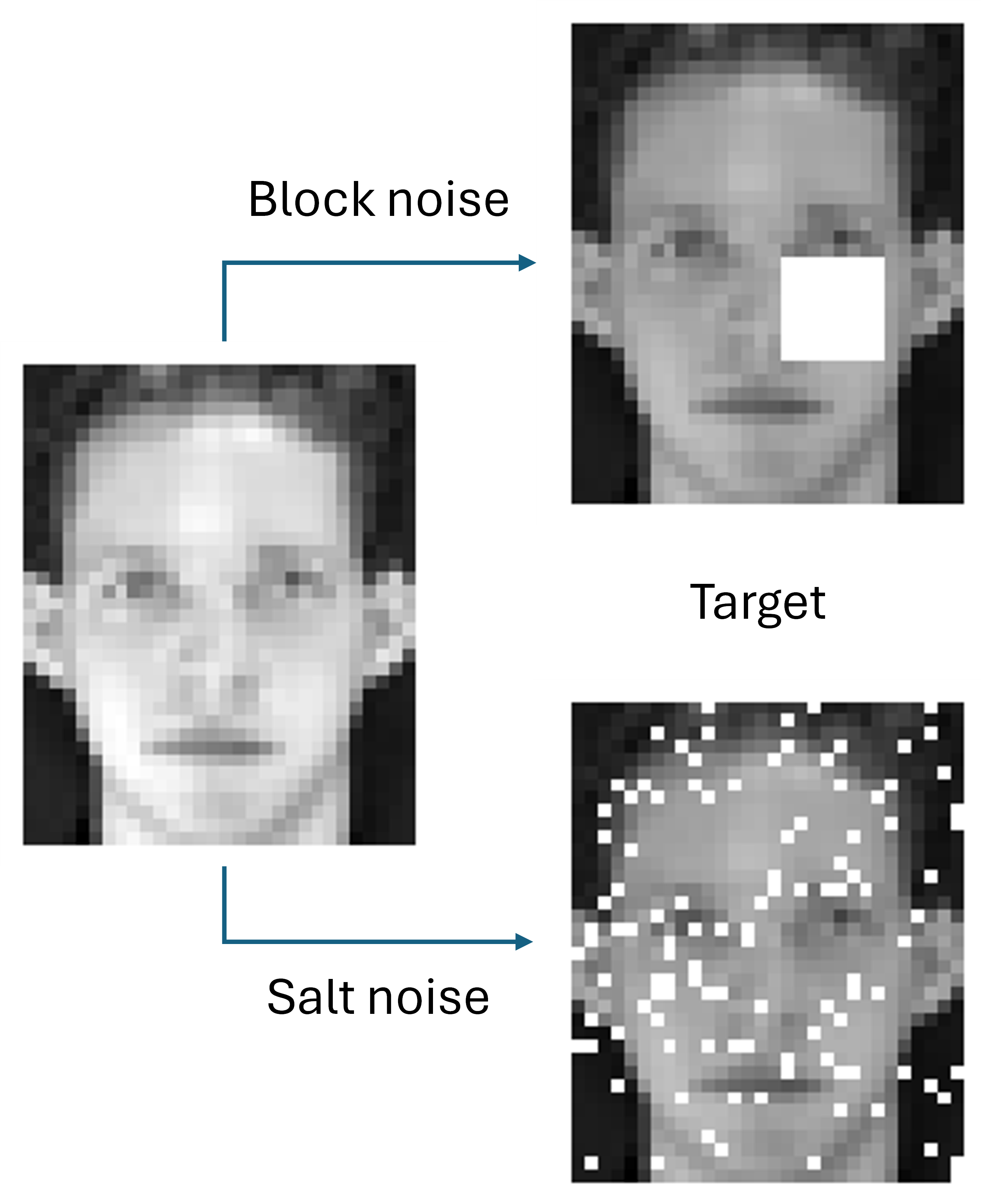}
    \caption{Sample from the ORL image database. Right panel shows corrupted images.}
    \label{fig:ORL-image_1}
\end{figure}

We compared our new approach with state-of-the-art weighted NMF approaches: CIM-NMF and Huber-NMF \cite{Du2012RobustNM}, using the \href{https://www.kaggle.com/datasets/kasikrit/att-database-of-faces}{ORL} (Olivetti Research Laboratory) and the \href{https://datarepo.eng.ucsd.edu/visiondata/iskwak/public_html/ExtYaleDatabase/ExtYaleB.html}{CroppedYaleb} datasets, which contain face images from 40 and 28 subjects, respectively, with variations in lighting and facial expressions. "Block" corruption (a randomly placed white rectangle) and "salt" corruption (randomly distributed white pixels ) were applied to the images (Figure \ref{fig:ORL-image_1}).
\subsection{Methods}
\subsubsection{Outline}
Before we discuss our novel approach, it is illuminating to consider this brief parable: A fisherman throws a rope to the pier and pulls the ship closer. A child, observing this, remarks to his father, ”Look at how strong the fisherman is; he is pulling the pier closer to the ship!”. So, we took the child’s remark seriously and gradually polished the data to make it more amenable to fast factorization. How does an iteration of the ”Target Polish” approach work? As with weighted least squares NMF, we start by computing the squared differences between the data points and the values corresponding to the current factorization. From this, we can derive weights using any of the weighting schemes already proposed in the literature. We then compute the polished target for each data point as a weighted average between the global median of the dataset and the original value. This approach nudges poorly approximated points toward the global median, while leaving well-approximated points largely unaffected. The global median is used in place of the global mean due to its greater robustness to outliers. Factors can now be updated based on the polished target using Fast-HALS rules. Importantly, to save time, the polished target is  updated further using an iteration step that depends on the relative distance between the previous and current polished targets. Once this iterative process is completed, a few iterations of Weighted NMF are performed to get the factorized data closer to the original data. 

\subsubsection{Mathematical formulation}
For clarity, the formulas are presented in the context of NMF, but they can be readily extended to NTF. \newline

The polished target \(\tilde{X}\) is defined as:
\begin{equation} \label{eq:dynamic_x}
    \tilde{X}_{ij}= (1-G_{ij})\text{med}(X)+G_{ij}X_{ij}
\end{equation}
where the weight function \(G_{ij}\), e.g. the one defined in equation \ref{eq:cim} or equation \ref{eq:huber}, is chosen for specific robustness purposes. Note that \(G_{ij}\) is determined by using the Frobenius distance between the current factorization and \(X\), rather than between the current factorization and \(\tilde{X}\). Next, the factorization error is minimized against \(\tilde{X}\) instead of \(X\), using the adaptive optimization criterion:
\begin{equation} \label{eq:dynamic_loss}
    \tilde{J} = \sum_{i,j} \left(\tilde{X}_{ij} - (WH^T)_{ij} \right)^2.
\end{equation}
Importantly, to ensure computational efficiency,  \(\tilde{X}\) is not polished again after every iteration, as its updating process requires modifying the weighting matrix, which depends on the time-consuming calculation of the error matrix for \(X\). Instead, the update frequency is based on the relative change in a small sample of the polished target between the current iteration, denoted as \(\tilde{X}_{current} \) , and the previous update, denoted as \(\tilde{X}_{ref}\) :

\begin{equation} \label{eq:relative_change_target}
    \text{Relative Change} = \frac{\sum_{i,j} \left( \tilde{X}_{{current}_{ij}} - \tilde{X}_{{ref}_{ij}} \right)^2}{\sum_{i,j} \left(\tilde{X}_{{current}_{ij}}\right)^2}.
\end{equation}

More specifically, the rows and columns of the polished target are sampled at a given step size, which is defined as follows:

\begin{equation} \label{eq:step_size}
    step\_size = { \left(\text{size}(X) \times fraction\right)}^{1/\text{dim}(X)}
\end{equation}

where \(size(X)=n \times m\) represents the total number of entries in \(X\) and \(\text{dim}(X)\) is the number of dimensions in \(X\). In our context, \(\text{dim}(X)\) is 2. This ensures that indices \(i\) and \(j\) in equation \ref{eq:relative_change_target} are sampled so that only a small portion of the polished target is updated with each iteration. The fully polished target is updated only if the relative change exceeds a defined threshold. Based on our experience with corrupted images, we set the fraction parameter to 0.001 and the relative change threshold to 0.05. \newline

As long as the fully polished target is not updated, Fast-HALS updates steadily decrease the optimization criterion \(\tilde{J}\), leveraging its convergence properties. Once the fully polished target is updated, however, \(\tilde{J}\) generally undergoes a further reduction, as illustrated in the following heuristic proof. \newline

For a given pair \((i,j)\), consider two extreme outcomes at the moment the polished target is updated:
\begin{enumerate}
    \item Poor approximation (\( G_{ij} \approx 0\))
    \item Highly accurate approximation (\( G_{ij} \approx 1\))
\end{enumerate}

We derive from equation \ref{eq:dynamic_loss} the following expressions for each case:
\begin{enumerate}
    \item  \(\tilde{J}_{ij}  \approx \left(\text{med}(X) - (WH^T)_{ij}\right)^2\)
    \item \(\tilde{J}_{ij}  \approx \left(X_{ij} - (WH^T)_{ij}\right)^2\)
\end{enumerate}

where \(\tilde{J}_{ij}\) represents the component of \(\tilde{J}\) for the pair \((i,j)\). \newline

\begin{figure}[H]
    \centering
    \includegraphics[width=0.8\linewidth]{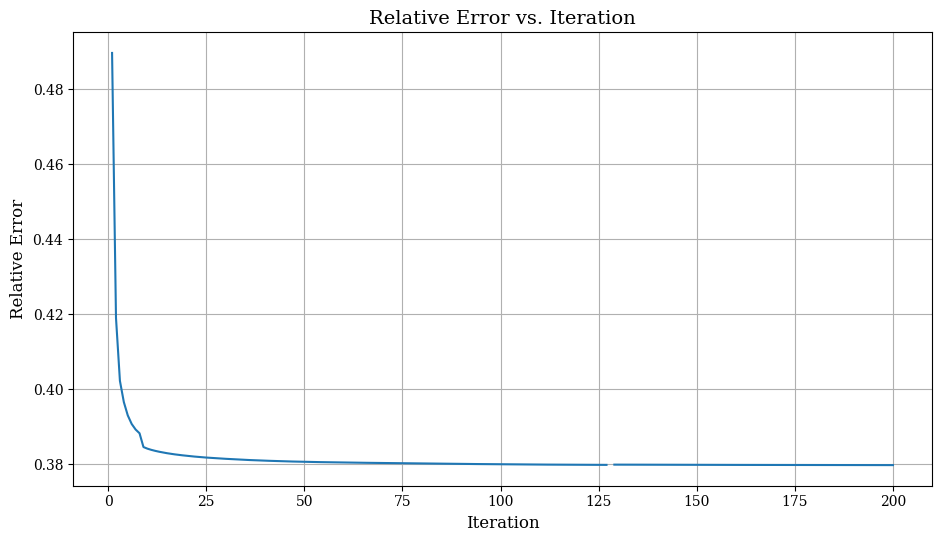}
    \caption{Relative error as a function of the update iteration}
    \label{fig:relative_errors}
\end{figure}

These equations illustrate how the optimization criterion is further minimized depending on approximation quality: When approximation quality is poor, \(X_{ij}\) is likely considered an outlier, making its replacement with \(\text{med}(X)\) a strategy that can further minimize the optimization criterion. This is because the matrices \(W\)and \(H\), derived through alternating projections, incorporate the full structure of \(X\) rather than relying solely on individual entries like \(X_{ij}\). Consequently, the approximation values \( (WH^T)_{ij}\) tend to gravitate toward the median of \(X\) rather than to extreme outliers.  Figure \ref{fig:relative_errors} shows how the error changes with each update, with the error decreasing smoothly, indicating that the target has been polished precisely as needed.

Conversely, if the approximation is nearly perfect, retaining \(X_{ij}\) in equation \ref{eq:dynamic_loss}  ensures the lowest value for the optimization criterion. \newline

\begin{figure}[H]
    \centering
    \includegraphics[width=0.5\linewidth]{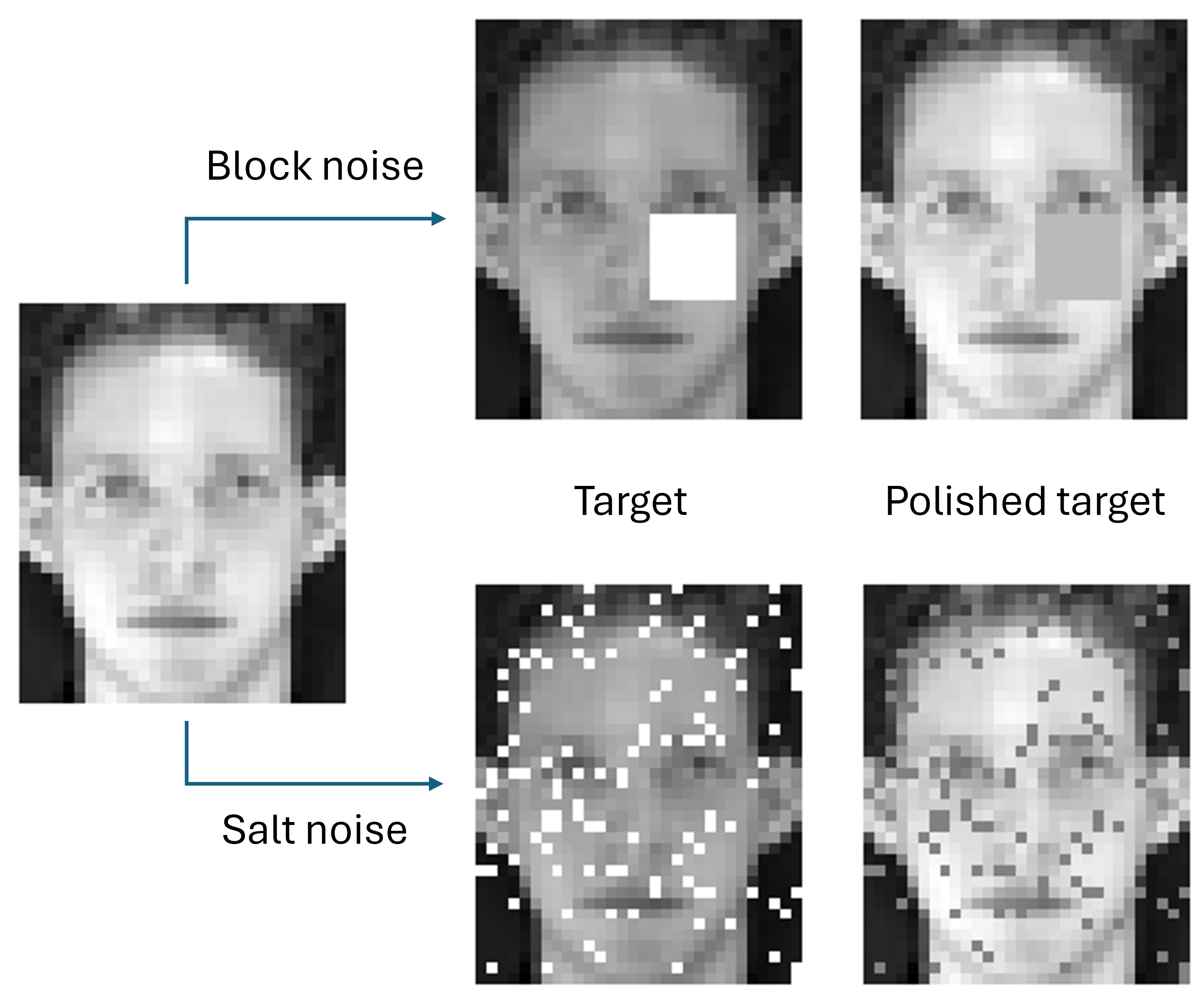}
    \caption{Sample from the ORL image database. Right panel shows the polished targets after convergence.}
    \label{fig:ORL-image_2}
\end{figure}

\begin{table*}
    \centering
    \resizebox{\textwidth}{!}{ % Adjust width
    \tiny 
    \begin{tabular}{lllrccc}
        \hline
        \textbf{Noise type} & \textbf{Weight} & \textbf{Method} & \textbf{RRE} & \textbf{ACC} & \textbf{NMI} & \textbf{Time (sec)} \\
        \hline
        \multirow{6}{*}{BLOCK} & \multirow{2}{*}{None} & Multiplicative updates& 0.4138 & 0.1973 & 0.3532 & 11.45\\
              &      & Fast-HALS& \textbf{0.4137} & \textbf{0.2190}  & \textbf{0.3845} & \textbf{0.66}\\
              \cline{2-7}
              & \multirow{2}{*}{CIM}  & Weighted NMF
& 0.3095 & 0.4175 & 0.5811 & 19.38\\
              &      & Target Polish& \textbf{0.1670}& \textbf{0.6970}& \textbf{0.8270}& \textbf{2.48}\\
              \cline{2-7}
              & \multirow{2}{*}{Huber} & Weighted NMF
& 0.4083 & 0.2135 & 0.3665 & 40.00\\
              &      & Target Polish& \textbf{0.3473}& \textbf{0.2997}& \textbf{0.4633}& \textbf{2.86}\\
              \hline
        \multirow{6}{*}{SALT} & \multirow{2}{*}{None} & Multiplicative updates
& \textbf{0.2879}& \textbf{0.5165}& \textbf{0.6740}& 10.81\\
              &      & Fast-HALS& 0.2883& 0.5150& 0.6731& \textbf{0.74}\\
              \cline{2-7}
              & \multirow{2}{*}{CIM}  & Weighted NMF
& \textbf{0.1365}& 0.6985& \textbf{0.8357}& 9.92\\
              &      & Target Polish& 0.1945& \textbf{0.6992}& 0.8326& \textbf{1.30}\\
              \cline{2-7}
              & \multirow{2}{*}{Huber} & Weighted NMF
& \textbf{0.1432}& \textbf{0.7170}& \textbf{0.8421}& 38.68\\
              &      & Target Polish& 0.2041& 0.6700& 0.8091& \textbf{1.45}\\
        \hline
    \end{tabular}
    }
    \caption{ORL image database}
    \label{tab:ORL_table}
\end{table*}

\begin{table*}
    \centering
    \resizebox{\textwidth}{!}{ % Adjust width
    \tiny 
    \begin{tabular}{lllrccc}
        \hline
        \textbf{Noise type} & \textbf{Weight} & \textbf{Method} & \textbf{RRE} & \textbf{ACC} & \textbf{NMI} & \textbf{Time (sec)} \\
        \hline
        \multirow{6}{*}{BLOCK} & \multirow{2}{*}{None} & Multiplicative updates
& 0.5844& 0.1766& 0.2820& 37.01\\
              &      & Fast-HALS& \textbf{0.5844}& \textbf{0.1780}& \textbf{0.2865}& \textbf{1.56}\\
              \cline{2-7}
              & \multirow{2}{*}{CIM}  & Weighted NMF
& \textbf{0.2328}& \textbf{0.2968}& \textbf{0.4025}& 22.33\\
              &      & Target Polish& 0.2632& 0.2756& 0.3860& \textbf{4.28}\\
              \cline{2-7}
              & \multirow{2}{*}{Huber} & Weighted NMF
& 0.5615& 0.2030& 0.3153& 84.39\\
              &      & Target Polish& \textbf{0.4502}& \textbf{0.2186}& \textbf{0.3388}& \textbf{10.42}\\
              \hline
        \multirow{6}{*}{SALT} & \multirow{2}{*}{None} & Multiplicative updates
& \textbf{0.4497}& 0.2090& \textbf{0.3402}& 23.49\\
              &      & Fast-HALS& 0.4510& \textbf{0.2112}& 0.3367& \textbf{1.36}\\
              \cline{2-7}
              & \multirow{2}{*}{CIM}  & Weighted NMF
& \textbf{0.2050}& 0.3018& 0.4120& 16.29\\
              &      & Target Polish& 0.2185& \textbf{0.3222}& \textbf{0.4325}& \textbf{4.22}\\
              \cline{2-7}
              & \multirow{2}{*}{Huber} & Weighted NMF
& \textbf{0.2018}& \textbf{0.3341}& \textbf{0.4490}& 83.51\\
              &      & Target Polish& 0.2602& 0.2677& 0.3892& \textbf{3.05}\\
        \hline
    \end{tabular}
    }
    \caption{CroppedYaleb image database}
    \label{tab:EYB_table}
\end{table*}

Finally, after convergence, it is \(\tilde{X}\) that has been factorized, rather than the original \(X\), as illustrated in Figure \ref{fig:ORL-image_2}. To factorize \(X\), the Target Polish factorization serves as the initialization for Weighted NMF. Importantly:
\begin{itemize}
    \item The final weight system obtained at the end of the Target Polish iterations is used throughout Weighted NMF.
    \item The optimization criterion that determines convergence is estimated using the same portion of the matrix defined by the sampling parameter \(step\_size\) used in equation \ref{eq:step_size}.
\end{itemize}
This approach ensures minimal Weighted NMF iterations and minimal impact on computational efficiency.

\subsection{Data and Code availability}
The Python \href{https://github.com/jasoncoding13/nmf/tree/master} {jasoncoding13} code was used to run Weighted NMF. The image databases we used can also be found in this repository.
The Python \href{https://github.com/Advestis/enAInem}{enAInem} code (\href{https://github.com/Advestis/enAInem}{GitHub - Advestis/enAInem: Algorithms for decomposing nonnegative multiway array and multi-view data into rank-1 nonnegative tensors}) was used to run NMF with the Target Polish approach.

\section{Results}

\subsection{Performance Metrics}
The \textit{Relative Reconstruction Error (RRE)}, defined as the ratio of root squared error to root sum of squares, was computed. Importantly, this computation uses the original—uncorrupted—matrix for error assessment. Given that each dataset includes variations in lighting and facial expressions for each individual, the performance in identifying the individual corresponding to a specific facial image (represented as a column of \(X\)) was evaluated using the following metrics:
\begin{itemize}
    \item \textit{Accuracy (ACC)}—the proportion of correct predictions relative to the total predictions made.
    \item \textit{Normalized Mutual Information (NMI)}—a measure of similarity between two clustering assignments.
\end{itemize}
It should be noted that ACC and NMI do not inherently designate one clustering assignment as the gold standard. \newline

The evaluation was conducted ten times for each method and type of noise. The small number of iterations is due to the poor performance of Weighted NMF. Corrupted pixels were randomly reassigned in each simulation. Since the results were very similar across the simulations, only the average performance and computational time are reported. We also evaluated standard NMF algorithms using multiplicative updates or Fast-HALS. 

\subsection{Results}

For the ORL data, the Target Polish approach consistently outperforms Weighted NMF when images are corrupted by block noise. The performance gap observed can be particularly significant for ACC, NMI and computational time. When images are corrupted by salt, the results are mixed, with a very small difference between the two approaches, except in computation time which is significantly lower with the Target Polish approach (Table \ref{tab:ORL_table}). \newline

For the CroppedYaleb dataset, the results are mixed, with a very small difference between the two approaches, except in computation time which is significantly lower with the Target Polish approach (Table \ref{tab:EYB_table}).\newline

\begin{figure}
    \centering
    \includegraphics[width=0.8\linewidth]{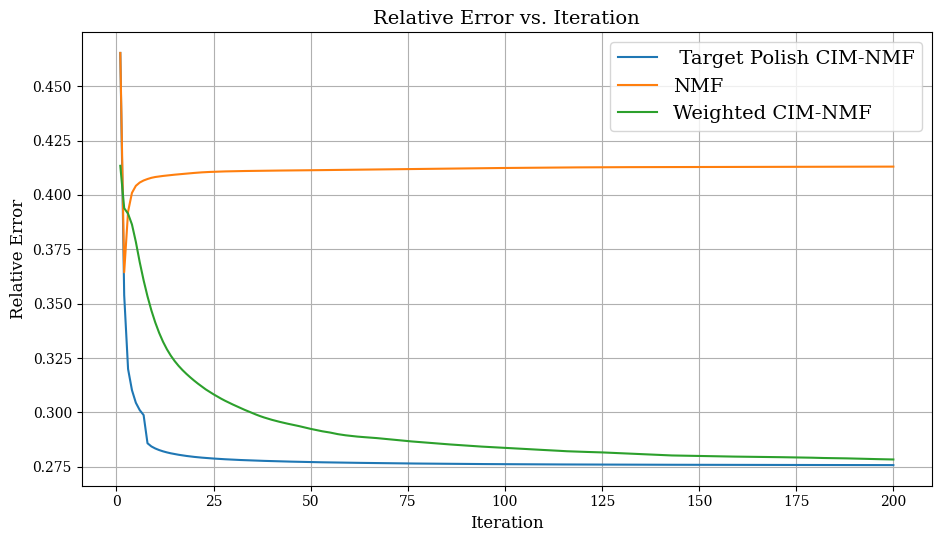}
    \caption{Relative error (using the non-corrupted data) as a function of the iteration}
    \label{fig:relative_error_non_corrupted}
\end{figure}

To assess the impact of applying the Target Polish approach with CIM weights, we further analyzed the progression of the relative error computed with respect to the original, uncorrupted images  (Figure \ref{fig:relative_error_non_corrupted}). The original images were artificially corrupted using block noise. The first 200 iterations were performed (for the Weighted CIM-NMF, each iteration actually consists of ten multiplicative updates). Notably, the Target Polish approach resulted in a much more pronounced decline in error during the first iterations than the Weighted NMF approach. Standard NMF initially showed a rapid error reduction, followed by a rebound to a plateau. This suggests that standard NMF is  affected more by corrupted images,  being misled by block corruption.

\section{Discussion}

This study focuses on enhancing the robustness of the Fast-HALS algorithm while leveraging its computational efficiency. In practical applications, our approach ensures superior computational performance that significantly outperforms state-of-the-art Weighted NMF, all while achieving comparable resistance to outliers. Our methodology notably extends beyond the conventional two-dimensional framework by seamlessly integrating with multidimensional arrays of dimensions greater than two. The Target Polish approach for multidimensional arrays of any dimension is provided by our enAInem code.\newline

Several key areas warrant further exploration.\newline

A fundamental priority is conducting a thorough examination of the convergence properties. This involves combining rigorous mathematical analysis with extensive simulation studies to validate the stability and effectiveness of our proposed method.\newline

One limitation of this study in image analysis is that it does not account for other types of outlier images, such as those with additive noise, structural anomalies, semantic deviations, or contrast irregularities. \newline

To enhance solution robustness, results obtained through different random initializations could be systematically integrated using the Integrated Sources Model (ISM) \cite{Fogel2024IntegratedSM}. \newline

Furthermore, this approach is likely to be extended to \textit{generalized} NMF or NTF frameworks. This generalization, which has been proposed in \cite{Ho2008}, can incorporate alternative optimization criteria, such as Kullback–Leibler (KL) divergence, by leveraging iteratively reweighted least squares (IRLS) \cite{Hampel1986robust}. Replacing IRLS with our Target Polish approach is expected to significantly increase its computational performance. \newline

In conclusion, the Target Polish approach offers a computationally efficient and robust factorization approach that strikes a balance between accuracy, speed, and resistance to outliers. Furthermore, this methodology has the potential to strengthen other optimization algorithms, offering new perspectives on enhancing computational stability across various factorization techniques.

%\bibliography{main}

\end{document}